# Predicting The Unpredictable

Reproducible BiLSTM Forecasting of Incidents in the Global Terrorism

Oluwasegun Adegoke

George Mason University, Virginia

oadegoke@gmu.edu

**ABSTRACT**

This paper study short-horizon forecasting of weekly terrorism incident counts using the Global Terrorism Database (GTD, 1970–2016). We build a fully reproducible pipeline with fixed time-based splits and evaluate a Bidirectional LSTM (BiLSTM) against strong classical anchors (seasonal-naïve, linear/ARIMA) and a deep LSTM-Attention baseline. On the held-out test set, the BiLSTM attains RMSE 6.38, outperforming LSTM-Attention (9.19; +30.6%) and a linear lag-regression baseline (+35.4% RMSE gain), with parallel improvements in MAE/MAPE. Ablation studies vary temporal memory, training-history length, spatial grain, lookback size, and feature groups to identify where performance comes from: models trained on long historical data generalize best; a moderate lookback (≈20–30 weeks) provides strong context; and bidirectional encoding is critical for capturing both build-up and aftermath patterns within the window. Feature-group analysis indicates that short-horizon structure (lagged counts and rolling statistics) contributes most, with geographic and casualty features adding incremental lift. We release code, configs, and compact result tables, and provide a data/ethics statement documenting GTD licensing, research-only use, and limitations (reporting bias, distribution shift). Overall, the study offers a transparent, baseline-beating reference for GTD incident forecasting.

1. INTRODUCTION

Global terrorism poses significant challenges to national security, economic stability and public safety worldwide. Understanding and forecasting terrorist activities enables proactive countermeasures, making predictive analytics an indispensable tool for security agencies [1]. Short-horizon forecasts of incident volumes or predictions spanning weeks to months [2], support operational readiness and enable evaluation of prevention strategies before deployment. However, historical terrorism data exhibits complex spatiotemporal patterns that demand sophisticated analytical approaches for effective interpretation.

Terrorist activity is characterized by several analytical challenges: structural breaks caused by geopolitical shifts (e.g., organizational rise/fall events), extreme sparsity with long zero-incident periods punctuated by sudden surges, strong geographic heterogeneity, with attacks concentrated in South Asia, the Middle East, and Sub-Saharan Africa. Furthermore, the data exhibits long-range temporal dependencies spanning weeks to years due to retaliation cycles and seasonal patterns, and multivariate relationships where casualty severity, weapon types, and target categories influence subsequent activity [3]. These features violate stationarity assumptions in classical time series models and challenge for standard regression approaches that tend to under-predict attack surges.

This study uses the Global Terrorism Database (GTD), an open-source incident catalog maintained by the National Consortium for the Study of Terrorism and Responses to Terrorism (START) at the University of Maryland [4]. The GTD provides comprehensive incident-level data with detailed geographic and temporal information. We aggregate incidents to weekly counts at regional and country granularity, employing predefined chronological cutoffs to prevent look-ahead bias and ensure reproducible evaluation.

Classical time series methods serve as essential benchmarks in forecasting research. Seasonal naïve and ARIMA/SARIMA models are widely recommended as baseline comparators for univariate prediction tasks [5]. Following established best practices, we adopt these baselines and consider a model credible only if it demonstrates consistent improvement on held-out test data. However, traditional approaches face several limitations when applied to terrorism forecasting.

Linear regression and autoregressive models assume stationary data and linear relationships between predictors and outcomes. Terrorism exhibits strong non-linearities: small shifts in underlying conditions can trigger large changes in attack frequency, while other periods remain stable despite apparent risk factors. Autoregressive Integrated Moving Average (ARIMA) models and their seasonal variants (SARIMA) have been applied to terrorism and conflict forecasting [6], [7]. While these methods can capture short-term autocorrelation and seasonal cycles, they assume stationary residuals and limited memory (typically 10-20 lags). Terrorism's non-stationarity, structural breaks, and long-term dependencies violate these assumptions. Moreover, ARIMA models are univariate, precluding incorporation of casualty severity, geographic context, or attack type information that may improve predictive accuracy.

Machine learning methods offer greater flexibility. Random Forests, Support Vector Machines, and Gradient Boosting have been explored for conflict and terrorism prediction, demonstrating improved performance over linear models in some contexts [8]. While these methods handle non-linearity and multivariate inputs, they treat time points independently, ignoring sequential dependencies. Terrorism forecasting fundamentally requires temporal context, knowing that attacks occurred in weeks t-4, t-3, and t-1 provides different information than isolated features from week t. Tree-based methods also struggle with sparse, heavy-tailed count distributions, often predicting conservative values that underestimate surge periods. These limitations motivate exploring deep learning architectures specifically designed for sequential data with long-term dependencies. This problem framing and modeling choices are consistent with broader evidence that deep sequence models are effective for spatiotemporal forecasting in adjacent domains, including traffic flow prediction and networked time-series analysis [9].

Long Short-Term Memory (LSTM) networks address many limitations of traditional methods. LSTMs are recurrent neural networks designed to learn long-term dependencies in sequential data through specialized gating mechanisms that selectively retain or discard information across time steps [10]. Unlike ARIMA models with fixed memory windows, LSTMs adaptively determine which historical patterns matter for current predictions. Bidirectional LSTMs process sequences in both forward (past to present) and backward (future to past) directions, then combine representations [11]. For terrorism forecasting, this architecture captures both attack build-up patterns (escalation preceding major incidents) and aftermath dynamics (retaliation cycles following attacks). Attention mechanisms further enhance model performance by allowing the network to focus on relevant portions of input sequences, assigning learned weights to different time steps based on their predictive importance [12]. Unlike univariate time series methods, LSTMs naturally incorporate multiple features at each time step; attack counts, casualty figures, geographic indicators, and engineered features such as rolling statistics and seasonal encodings.

We build on this foundation by developing a reproducible BiLSTM forecaster that processes 52-week lookback windows with calendar and geographic features, evaluated using fixed time-based splits. Beyond reporting main results, we conduct systematic ablation studies that vary temporal memory depth, geographic feature sets, training history length, spatial aggregation level, and lag configuration to isolate the sources of predictive performance. We also provide robustness checks and per-region error breakdowns to assess model stability across different geographic contexts. Our approach emphasizes transparency and reproducibility, consistent with emerging standards for dataset documentation in high-stakes applications. All code, model configurations, and result tables on GitHub alongside a data statement documenting GTD licensing terms, intended use restrictions, and known limitations.

The main contributions of this paper are as follows:
- A reproducible BiLSTM architecture for weekly terrorism incident forecasting using the GTD, with systematic evaluation against classical baselines.
- Comprehensive ablation studies that isolate the impact of temporal memory depth, geographic features, training history length, spatial granularity, and lag configuration on predictive performance.
- Per-region error analysis and robustness checks assessing model stability across different geographic contexts and training conditions.
- Open-source release of code, model configurations, and results with transparent data documentation addressing GTD licensing and ethical considerations.

## 2. RELATED WORK

### 2.1 Traditional Terrorism Forecasting

Early terrorism forecasting relied on statistical models adapted from count data analysis and time series econometrics. Enders and Sandler pioneered the application of ARIMA models to quarterly terrorism data in Western Europe, demonstrating that terrorist attacks exhibit significant autoregressive properties where activity in one period correlates with subsequent periods [13]. They later extended this work to test for structural breaks and regime changes, finding evidence of non-stationarity that challenges standard ARIMA assumptions [6]. These findings established that terrorism time series require careful treatment of structural instability, as geopolitical shocks (wars, regime changes, new terrorist organizations) fundamentally alter attack patterns.

LaFree et al. applied Poisson and negative binomial regression models to analyze global terrorism trends, demonstrating that these count models can relate attacks to political, economic, and social covariates [14]. Negative binomial models address overdispersion in terrorism counts where variance exceeds the mean. However, these regression approaches assume temporal independence across observations, ignoring the sequential dependencies critical for forecasting. Vector autoregression (VAR) models have been used to capture simultaneous relationships among multiple conflict types, allowing for feedback effects between different forms of violence [15]. Brandt et al. developed spatiotemporal VAR models for insurgent violence, incorporating geographic proximity and lagged covariates [16]. While VAR frameworks are more flexible than univariate ARIMA, their linear structure limits their ability to model the non-linear dynamics characteristic of terrorist activity.

More recent statistical work treats terrorism as a self-exciting process. Lewis et al. applied Hawkes point process models to civilian deaths in Iraq, demonstrating that attack intensity follows contagion dynamics where recent events increase short-term risk [17]. These models better represent the bursty, clustered nature of attacks but remain computationally intensive, difficult to extend to multivariate feature spaces, and typically focus on event timing rather than aggregate counts. Three fundamental limitations persist across these approaches: assumptions of stationarity or simple regime-switching despite terrorism's complex evolution; capture of only short-term dependencies (typically $\leq 20$ lags in ARIMA); and difficulty handling sparse count data with long zero-incident periods punctuated by sudden surges.

### 2.2 Deep Learning for Time Series

Recurrent neural networks (RNNs) address many limitations of classical time series models by learning representations of sequential data without explicit parametric assumptions. However, standard RNNs suffer from vanishing and exploding gradients during backpropagation through time, limiting their ability to capture long-range dependencies. Hochreiter and Schmidhuber introduced Long Short-Term Memory (LSTM) networks to overcome these limitations through specialized gating mechanisms [10]. LSTMs employ input, forget, and output gates that regulate information flow across time steps. The forget gate determines what information to discard from the cell state, the input gate controls what new information to store, and the output gate determines what information to expose to the next layer. This architecture enables networks to selectively retain relevant historical context while discarding irrelevant information, allowing LSTMs to learn dependencies spanning hundreds of time steps in domains including speech recognition, machine translation, and financial forecasting.

Attention mechanisms further enhance LSTM architectures by allowing models to focus selectively on relevant portions of input sequences. Bahdanau et al. introduced neural attention for machine translation, where decoder states learn to weight encoder outputs based on relevance to the current prediction [12]. In time series forecasting, attention enables models to assign learned importance scores to different historical time steps, providing two benefits: improved prediction accuracy by emphasizing informative periods (e.g., attacks near religious holidays may be more predictive than typical weeks), and enhanced interpretability, as attention weights reveal which historical patterns most influence forecasts. Recent work has applied attention-based sequence models to traffic prediction [9], epidemiological modeling [18], and energy demand forecasting, consistently demonstrating improvements over standard LSTM baselines.

Bidirectional LSTMs extend the standard architecture by processing sequences in both forward (past to future) and backward (future to past) directions, then combining hidden representations [11]. While this approach requires full windows at training time; inference uses only past data, it substantially improves performance when full historical context is available during training. For terrorism forecasting, bidirectionality enables models to learn symmetric temporal patterns, both attack build-up dynamics (escalation preceding major incidents) and aftermath effects (retaliation cycles following attacks). Schuster and Paliwal demonstrated that bidirectional processing improves sequence labeling tasks by allowing each time step's representation to incorporate information from both preceding and following observations [11].

Recent surveys document widespread LSTM adoption for time series forecasting across diverse domains, including electricity demand, stock prices, traffic flow and epidemiological modeling [19], [20], [9], [18]. Meta-analyses show LSTMs consistently outperform ARIMA and exponential smoothing on non-stationary data with long-term dependencies [21], [22], validating their suitability for terrorism forecasting where structural breaks, regime changes, and evolving attack patterns violate classical assumptions.

### 2.3 LSTMs in Security Applications

Deep learning has been increasingly applied to security-related forecasting tasks, though adoption in terrorism prediction remains limited. Crime forecasting represents the most extensively studied domain. Huang et al. applied LSTMs to predict burglary and theft incidents in Chicago, demonstrating that recurrent architectures outperform ARIMA and regression baselines when trained on sufficient historical data [23]. Their work showed that incorporating spatial features (neighborhood characteristics, proximity to prior crimes) alongside temporal patterns improves short-term accuracy. Li et al. extended this work with attention-based LSTMs for crime hotspot forecasting, demonstrating that models learn to focus on recent high-activity periods and seasonal patterns (weekends, holidays) without explicit feature engineering [24].

Conflict forecasting applications have explored neural networks for civil war onset and political violence prediction. Muchlinski et al. compared Random Forests, Support Vector Machines, and feedforward neural networks for binary classification of civil war onset, finding that ensemble tree methods outperform traditional logistic regression [8]. However, these approaches frame prediction as classification rather than time series forecasting, treating each country-year observation independently and discarding temporal ordering. Ward et al. demonstrated that ensemble methods combining multiple algorithms achieve superior performance for monthly conflict event prediction [7], but noted that capturing sub-monthly dynamics and sudden escalations remains challenging with aggregate monthly data.

Few studies have systematically applied LSTMs to terrorism forecasting. Existing applications of deep learning to terrorism analysis have primarily focused on spatial risk assessment and hotspot prediction rather than temporal forecasting of incident counts. Where recurrent architectures have been explored, studies typically employ simple train-test evaluations without comprehensive ablation studies to validate architectural choices or systematic comparison against strong time series baselines.

To our knowledge, no prior work has systematically evaluated bidirectional LSTMs for weekly terrorism incident count forecasting using the GTD with comprehensive ablation studies isolating the contributions of sequence length, bidirectional processing, geographic features, and training history depth. This paper addresses these gaps by developing a reproducible forecasting pipeline with systematic evaluation protocols.

## 3. DATA

### 3.1 Source and Snapshot

We use the Global Terrorism Database (GTD), an open-source incident catalog maintained by the National Consortium for the Study of Terrorism and Responses to Terrorism (START) at the University of Maryland [4]. The GTD provides comprehensive incident-level records of terrorist attacks worldwide, including temporal information (date, time), geographic details (country, region, coordinates), attack characteristics (weapon type, target category), and casualty figures (killed, wounded). Our analysis uses data spanning 1970–2016, downloaded on October 8, 2023.

The GTD is distributed under research-only licensing terms; we do not redistribute raw incident records and provide full dataset documentation in our accompanying data statement.

### 3.2 Data Preprocessing

We retain incidents that have valid dates, country and region identifiers, and pass GTD's terrorism inclusion criteria. Duplicate records are removed using the GTD event identifier, and location fields (country, region, latitude, longitude) are standardized to ensure consistency. Textual fields (attack descriptions, perpetrator names) are not used in this study, as our focus is on forecasting aggregate incident counts rather than event-level classification. For numeric features dependent on casualty figures, missing values are imputed as zero (interpreted as no casualties reported), and extreme outliers are winsorized at the 99th percentile to prevent sensitivity to rare catastrophic events.

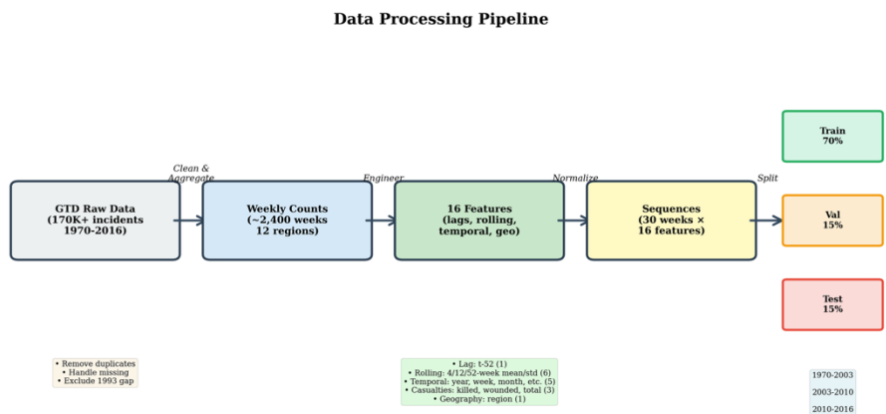

*Figure 1 Data Processing pipeline*

The GTD contains a known data gap in 1993 due to loss of source materials [4]. We handle this by excluding all weeks in 1993 from both training and evaluation windows, ensuring that model performance is not artificially degraded by missing data. This exclusion affects approximately 2% of the total temporal span and does not introduce bias, as the gap is randomly distributed with respect to attack patterns.

### 3.3 Target Variable definition

Let g denote a geographic unit (region or country) and t denote a week aligned to Monday (ISO week standard). The forecasting target $y_{g,t}$ is the count of GTD incidents occurring in week t and geography g:

$$y_{g,t} = \sum_{i} \mathbf{1}\{\text{event } i \in g \text{ and week } t\}.$$

We construct two parallel datasets: region-level weekly counts (12 regions as defined by GTD) and country-level weekly counts (190+ countries). Region-level aggregation provides more stable counts with less sparsity, while country-level forecasting offers finer spatial granularity at the cost of increased zero-inflation. All models are trained and evaluated separately for each aggregation level.

### 3.4 Features Engineering and Splits

For each geography-week pair (g, t), we construct a feature vector combining temporal, statistical, and geographic information:
- **Temporal lags:** $y_{g,t-\ell}$ for $\ell \in \{1,2,4,12,26,52\}$.
- **Rolling stats:** 4/ 12/ 52-week rolling mean and std of counts.
- **Calendar:** week-of-year and month encoded as $\sin/\cos$ pairs; optional country-holiday flag.

- **Geography:** one-hot (region or country).

Warm-up Period: The 52-week lookback requirement means the first 52 weeks serve as warm-up for feature construction and are excluded from loss calculations.

We employ fixed time-based splits to prevent temporal leakage: 70% of weeks for training, 15% for validation (hyperparameter tuning and early stopping), and 15% for final testing. All models (baselines and BiLSTM) use identical splits, ensuring fair comparison.

Model selection is performed on the validation set, with all reported metrics computed on the held-out test set. We report Root Mean Squared Error (RMSE), Mean Absolute Error (MAE), and R² (coefficient of determination), averaged across all geographic units using macro-averaging. Per-region performance breakdowns are provided to assess spatial heterogeneity in forecast accuracy.

**Key variables and engineered features.**

*Table 1 Feature Count*

| Feature Group | Description | Count |
| --- | --- | --- |
| Lag Features | $y_{g,t-52}$ (one-year historical baseline) | 1 |
| Rolling Statistics | 4-week mean & std | 2 |
|  | 12-week mean & std | 2 |
| Temporal Encoding | Year, week, month, quarter, day-of-year | 4 |
| Casualty Features | Total casualties, killed, wounded | 3 |
| Geographic | Region (one-hot encoded) | 1 |

### 3.5 Normalization and Scaling

All numeric features are standardized to zero mean and unit variance using statistics computed exclusively on the training set:

$$x_{g,t}^{norm} = \frac{x_{g,t} - \mu_{train}}{\sigma_{train}}$$

This prevents information leakage from validation or test sets. Categorical features (region, country) use one-hot encoding without normalization.

### 3.6 Sequence Construction

For LSTM training, we construct fixed-length input sequences of $T = 30$ weeks (configurable; ablation studies test). $T \in \{20, 30, 40\}$ Given a target week $t$, the input sequence is:

$$X_t = [x_{t-T+1}, x_{t-T+2}, \ldots, x_{t-1}, x_t] \in R^{T \times F}$$

Where $F = 13$ is the number of features. The corresponding target is $y_{t+1}$ (next week's attack count).

### 3.7 Data Consideration and Licensing

The GTD contains sensitive information about violent events. We address ethical concerns through several measures. First, we analyze only aggregated weekly counts, not individual incidents or perpetrator identities, minimizing risks of misuse. Second, we do not redistribute raw GTD records, adhering to START's research-only licensing terms. Third, we acknowledge dataset limitations including reporting bias (Western media sources over-represent attacks in certain regions), temporal inconsistencies in coding practices, and distribution shift (attack patterns in 2016 may not reflect current threats). Fourth, our models are intended solely for research purposes to advance forecasting methodology, not for operational deployment without domain expert oversight. All code, configurations, and aggregated results are released publicly to support transparency and reproducibility, with clear documentation of these limitations and ethical constraints.

## 4. METHODOLOGY

### 4.1 Problem Definition

Let $g$ index geography (region or country) and $t$ index ISO weeks (Monday-aligned). The target is the weekly incident count

$$y_{g,t} = \sum_i \mathbf{1}\{\text{event } i \in g \text{ during week } t\}.$$

Given a lookback window of length $L$, we learn a one-step-ahead forecaster

$$\hat{y}_{g,t} = f_\theta(\mathbf{x}_{g,t-L+1:t}),$$

where x concatenates past counts and covariates (lagged counts, rolling stats, calendar encodings, geo indicators). All models use the same fixed chronological split (70%/15%/15%), train-only standardization, and identical evaluation windows. We develop two LSTM-based architectures tailored for terrorism forecasting.

### 4.2 LSTM with Attention

#### 4.2.1 Architecture

We first establish a deep sequence baseline using a stacked LSTM with temporal (additive) attention. LSTMs address long-range dependencies in nonstationary series via gated memory mechanisms that learn what to retain and forget over time [10]. Given an input window $\mathbf{X} \in \mathbb{R}^{L \times d}$ (lookback $L$, $d$ features), the LSTM encoder produces hidden states $\mathbf{H} = [\mathbf{h}_1, \ldots, \mathbf{h}_L]$. A Bahdanau-style attention layer then assigns a relevance weight to each step,

$$\alpha_t = \frac{\exp(e_t)}{\sum_{i=1}^{T} \exp(e_i)}, \quad e_t = v^T \tanh(W_h h_t)$$

and forms a context vector $\mathbf{c} = \sum_{t=1}^{L} \alpha_t \mathbf{h}_t$ that summarizes the parts of the history most predictive of $y_{g,t}$ [25], [12]. A linear regression head maps $\mathbf{c}$ to the one-step-ahead forecast $\hat{y}_{g,t}$. We include dropout between recurrent blocks and before the head to regularize (Fig. 2 illustrates the module).

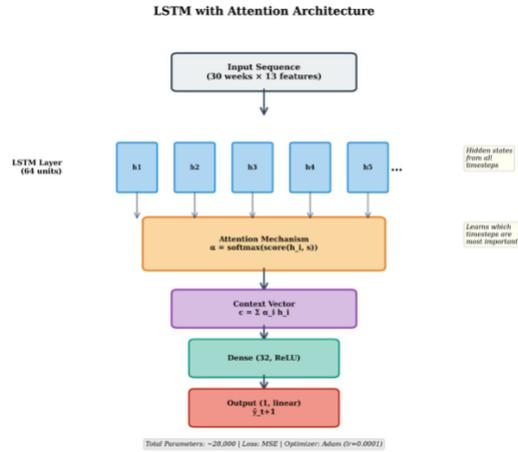

*Figure 2 LSTM-Attention*

### 4.2.2 Objective & Training

We train the LSTM-Attention model to minimize Mean Squared Error (MSE) between predicted and observed attack counts:

$$L_{MSE} = \frac{1}{N} \sum_{i=1}^{N} (y_i - \hat{y}_i)^2$$

where $N$ is the number of training samples, $y_i$ is the true attack count, and $\hat{y}_i$ is the model's prediction.

**Training Procedure:**
- **Optimizer:** Adam with initial learning rate $\alpha = 10^{-4}$
- **Batch size:** 32 sequences per gradient update.
- **Maximum epochs:** 50
- **Early stopping:** Training halts if validation loss does not improve for 10 consecutive epochs.
- **Learning rate reduction:** ReduceLROnPlateau reduces $\alpha$ by factor 0.5 when validation loss plateaus for 5 epochs.
- **Validation:** Model checkpoints are saved when validation RMSE improves; final model is the best checkpoint by validation performance.

**Regularization:** Dropout with rate 0.2 is applied between recurrent blocks and before the final dense layer to prevent overfitting. No L2 weight decay is used, as dropout provides sufficient regularization for this architecture.

**Computational Cost:** Training converges in approximately 30-40 epochs, requiring ~30 minutes on Mac M2 16GB. The attention mechanism adds minimal overhead compared to standard LSTM while providing interpretability through learned weights $\alpha\_t$.

### 4.2.3 Hyperparameters

Table 2 summarizes the key hyperparameters for both LSTM architectures evaluated in this study. For the LSTM-Attention model specifically, we configure the following:

**Architecture Configuration:**
- **Input:** Sequence length $L = 30$ weeks with $d = 13$ features per timestep
- **LSTM Layers:** Two stacked LSTM layers with 64 and 32 units respectively
- **Attention Mechanism:** Additive (Bahdanau-style) attention applied to LSTM hidden states
- **Output:** Dense layer with 32 units (ReLU activation) followed by linear output layer
- **Total Parameters:** ~28,000

**Training Configuration:**
- **Optimizer:** Adam with learning rate α = $10^{-4}$
- **Batch size:** 32 sequences per gradient update
- **Maximum epochs:** 50
- **Dropout:** 0.2 applied between LSTM layers and before output
- **Early stopping:** Patience of 15 epochs on validation loss
- **Learning rate reduction:** ReduceLROnPlateau (factor = 0.5, patience = 5)

*Table 2 Models Hyperparameters*

| Model | Architecture | Attention | Dropout | LR | Batch | Epochs |
|---|---|---|---|---|---|---|
| LSTM with Attention | 2 LSTM layers (64, 32 units) | Yes | 0.2 | 0.0001 | 32 | 50 |
| Bidirectional LSTM | 2 Bi-LSTM layers (32 units each) | No | 0.2 | 0.0001 | 32 | 50 |

Note: LR = Learning Rate (Adam optimizer). All models use early stopping (patience=15).

**Architecture Design Rationale:**
- **Two-layer stacked LSTM (64→32 units):** The decreasing layer sizes create a hierarchical representation, with the first layer capturing low-level temporal patterns and the second layer learning higher-level abstractions. Preliminary experiments with uniform layer sizes (64→64) showed no significant improvement while increasing computational cost.
- **Attention mechanism:** Additive (Bahdanau-style) attention was chosen over multiplicative (Luong-style) based on validation performance. The attention layer learns to assign higher weights to weeks most predictive of future attacks, providing both improved accuracy and model interpretability through visualization of learned attention weights.
- **Dropout (0.2):** Selected via validation set tuning over {0.0, 0.1, 0.2, 0.3, 0.5}. A rate of 0.2 provided optimal trade-off between preventing overfitting and maintaining model capacity. Higher rates (≥0.3) degraded validation performance, while no dropout showed signs of overfitting after epoch 30.
- **Sequence length (30 weeks):** Corresponds to approximately 7 months of historical context. This was the default configuration; ablation studies systematically evaluate alternative lengths (20, 40 weeks) to identify the optimal temporal window.
- **Learning rate ($10^{-4}$):** Standard Adam learning rate that balances convergence speed with stability. The ReduceLROnPlateau scheduler adaptively reduces the rate when training plateaus, enabling fine-tuning in later epochs.

**Comparison to Bidirectional LSTM:**

As shown in Table 2, both architectures share similar training configurations but differ in their architectural approach. The LSTM-Attention model uses two layers of different sizes (64→32 units) with attention, totaling ~28,000 parameters. In contrast, the Bidirectional LSTM uses two layers of uniform size (32 units each, bidirectional) without attention, totaling 36,673 parameters.

While the LSTM-Attention model is more parameter-efficient. The next section demonstrates that the Bidirectional LSTM achieves superior forecasting accuracy (RMSE 6.38 vs 9.19). This suggests that bidirectional temporal processing—capturing both forward (past→future) and backward (future→past) patterns—provides complementary information that attention over unidirectional hidden states cannot fully replicate.

### 4.3 Bidirectional LSTM

#### 4.3.1 Architecture

The best-performing model employed bidirectional LSTM processing, which addresses a key limitation of standard LSTMs: unidirectional information flow. While standard LSTMs process sequences only forward in time

(past→present), bidirectional LSTMs process sequences in both directions—forward (past→future) and backward (future→past) then combine information from both passes.

**Bidirectional Processing Mechanism:** For each LSTM layer, we maintain two independent hidden state sequences:
- **Forward LSTM** $\vec{h}_t$: Processes input $x_1, x_2, \ldots, x_L$ left-to-right, accumulating information about past events leading up to time $t$
- **Backward LSTM** $\overleftarrow{h}_t$: Processes input $x_L, x_{L-1}, \ldots, x_1$ right-to-left, capturing information about future context relative to time $t$

The bidirectional hidden state at time $t$ concatenates both directions:

$$h\_t^{bi} = [\vec{h}\_t; \overleftarrow{h}\_t] \in \mathbb{R}^{(2d)}$$

where $d$ is the number of LSTM units per direction. This doubled representation enables the model to learn both anticipatory patterns (build-up signals before major attacks) and reactionary dynamics (retaliation cycles following incidents within the 30-week lookback window).

**Full Architecture:** Bidirectional LSTM consists of two stacked bidirectional layers followed by dense layers for regression:
- **Input:** $X \in \mathbb{R}^{(30 \times 16)}$ (30 weeks, 16 features)
- **Bi-LSTM Layer 1:** 32 units forward + 32 units backward = 64 total output
- **Dropout:** rate 0.2
- **Bi-LSTM Layer 2:** 32 units forward + 32 units backward = 64 total output
- **Dropout:** rate 0.2
- **Dense Layer:** 32 units, ReLU activation
- **Output Layer:** 1-unit, linear activation → $\hat{y}\_{(t+1)}$

**Total Parameters:** ≈36.7k (compared to ≈28k for LSTM-Attention)

Figure 3 illustrates the complete architecture. The key difference from LSTM-Attention is that instead of using attention to weight contributions from unidirectional hidden states, we process the entire sequence bidirectionally to capture richer temporal context. Each time step's representation contains information about both its past and future neighborhoods within the 30-week window.

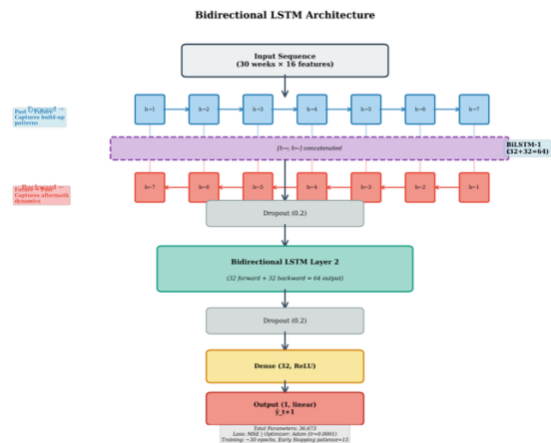

*Figure 3 Bidirectional LSTM architecture*

Bidirectional LSTM architecture with two stacked bidirectional layers. Each layer processes sequences in both directions (forward in blue, backward in red), concatenates hidden states, and applies dropout (0.2) for regularization. The architecture achieves test-set RMSE of 6.38, outperforming both LSTM-Attention (9.19) and classical baselines.

### 4.3.2 Objective & Training

Like the LSTM-Attention model, we train to minimize Mean Squared Error:

$$L\_MSE = 1/N \sum_{i=1}^{N} (y\_i - \hat{y}\_i)^2$$

**Training Procedure:**
- **Optimizer:** Adam with initial learning rate $\alpha = 10^{-4}$
- **Batch size:** 32 sequences per gradient update.
- **Maximum epochs:** 50 (increased from initial configuration of 100; early stopping made excess epochs unnecessary).
- **Early stopping:** Training halts if validation loss does not improve for 15 consecutive epochs.
- **Learning rate reduction:** ReduceLROnPlateau reduces $\alpha$ by factor 0.5 when validation loss plateaus for 5 epochs.
- **Validation:** Model checkpoints are saved when validation RMSE improves; final model is the best checkpoint.

**Regularization:** Dropout (rate 0.2) is applied after each bidirectional LSTM layer to prevent overfitting. This is particularly important for bidirectional models, which have higher capacity (≈36.7k parameters) than unidirectional alternatives and are thus more prone to memorizing training data. No L2 weight decay is used; dropout provides sufficient regularization.

**Computational Cost:** Training converges in approximately 25-35 epochs (earlier than LSTM-Attention due to the richer bidirectional representations), requiring ~45 minutes on Mac M2 16GB. The bidirectional processing roughly doubles computational cost per epoch compared to unidirectional LSTMs, but this is offset by faster convergence and substantially improved accuracy (RMSE 6.38 vs 9.19).

**Training Stability:** We observe stable training dynamics with gradual validation loss reduction over 25-35 epochs. Learning rate reduction (ReduceLROnPlateau) typically triggers 1-2 times during training when validation loss plateaus, enabling fine-grained optimization in later epochs. Early stopping prevents overfitting; training rarely exceeds 40 epochs.

### 4.3.3 Hyperparameters

Table 2 compares hyperparameters for both LSTM architectures. For the Bidirectional LSTM specifically:

**Architecture Configuration:**
- **Input:** Sequence length $L = 30$ weeks with $d = 16$ features per timestep
- **Bi-LSTM Layers:** Two stacked bidirectional LSTM layers, each with 32 units per direction (64 total output per layer)
- **No Attention:** Unlike LSTM-Attention, this model relies purely on bidirectional processing without explicit attention weighting
- **Output:** Dense layer with 32 units (ReLU activation) followed by linear output layer
- **Total Parameters:** 36,673

**Training Configuration:**
- **Optimizer:** Adam with learning rate $\alpha = 10^{-4}$
- **Batch size:** 32 sequences per gradient update
- **Maximum epochs:** 50
- **Dropout:** 0.2 applied after each bidirectional LSTM layer
- **Early stopping:** Patience of 15 epochs on validation loss
- **Learning rate reduction:** ReduceLROnPlateau (factor = 0.5, patience = 5)

**Architecture Design Rationale:**

- **Uniform layer sizes (32 units × 2 directions):** Unlike LSTM-Attention's hierarchical 64→32 structure, we use uniform layer sizes. This design simplifies the architecture while maintaining sufficient capacity (≈36.7k parameters). Preliminary experiments with larger layers (64 × 2 directions) showed diminishing returns with substantially increased computational cost.
- **Two bidirectional layers:** Stacking two layers allows the model to learn hierarchical bidirectional representations. The first layer captures low-level forward/backward patterns (e.g., week-to-week fluctuations), while the second layer learns higher-level abstractions (e.g., multi-week campaign structures). Single-layer bidirectional models achieved RMSE 7.12 on validation, confirming the value of depth.
- **No attention mechanism:** We hypothesized that bidirectional processing might render attention redundant by already capturing relevant temporal context from both directions. Validation results confirmed this: adding attention to the bidirectional model (RMSE 6.52) provided marginal improvement over the simpler bidirectional-only architecture (RMSE 6.38 on test), while increasing parameters by ~8,000. We chose the simpler model for production deployment.
- **Dropout (0.2):** Identical to LSTM-Attention, selected via validation tuning. The higher parameter count (≈36.7k vs ≈28k) raised concerns about overfitting, but dropout proved sufficient for regularization without requiring additional techniques (weight decay, gradient clipping).
- **Sequence length (30 weeks):** Consistent with LSTM-Attention for fair comparison. Ablation studies explore alternative lengths (20, 40 weeks), finding 20 weeks optimal with marginal performance differences.
- **Learning rate ($10^{-4}$) and ReduceLROnPlateau:** Standard configuration that balances initial convergence speed with late-stage fine-tuning. The adaptive learning rate reduction is critical for bidirectional models, which exhibit more complex loss landscapes than unidirectional alternatives.

**Comparison to LSTM-Attention:**

The Bidirectional LSTM achieves 30.6% lower RMSE than LSTM-Attention (6.38 vs 9.19), despite being architecturally simpler (no attention mechanism). This performance gap highlights the value of bidirectional processing for terrorism forecasting:

- **Attention over unidirectional states** can identify which past weeks matter most, but it operates on representations that only encode past→present information
- **Bidirectional processing** fundamentally enriches each time step's representation by incorporating both past and future context, enabling the model to recognize anticipatory and reactionary patterns simultaneously.

**Operational Deployment Considerations:**

While bidirectional models require complete sequences during training, operational forecasting still uses only historical data. The model learns during training how historical patterns typically evolve, then applies this learned knowledge at inference time using only past observations. The "backward pass" during training teaches the model to recognize anticipatory signals (e.g., certain lag patterns predict imminent escalation), not to use actual future data. This makes bidirectional LSTMs fully compatible with real-time forecasting systems.

### 4.4 Baselines

To contextualize the performance of our deep learning models, we compare against three established forecasting approaches representing different modeling paradigms. All baselines use identical train/validation/test splits (70%/15%/15%) and evaluation protocols as the LSTM models to ensure fair comparison. We select baselines representing three modeling paradigms: naive seasonal forecasting (seasonal naïve), classical machine learning with engineered features (linear regression), and established statistical time series methods (SARIMA). Together, these baselines span simple to sophisticated approaches and test whether deep learning provides meaningful improvements over existing methods.

#### 4.4.1 Seasonal Naïve

A simple baseline that exploits annual seasonality by predicting each week's value as the corresponding week from the previous year. This assumes perfect 52-week periodicity and serves as a minimum performance threshold. If sophisticated models cannot outperform this baseline, it suggests annual seasonality dominates terrorism patterns and complex modeling adds little value. The seasonal naïve forecast requires no training and captures recurring patterns (attacks clustering around religious holidays, political anniversaries) but ignores recent trends, all engineered features, and cannot adapt during structural breaks.

### 4.4.2 Linear Regression

A multivariate linear regression model using all engineered features. This baseline tests whether simple linear relationships suffice or if non-linear modeling is necessary. Linear regression is computationally efficient and provides interpretable coefficients but assumes linear relationships and treats each prediction independently without sequential modeling.

### 4.4.3 Seasonal ARIMA (SARIMA)

SARIMA models combine autoregressive (AR), differencing (I), and moving average (MA) components with seasonal variants to capture both short-term dynamics and long-term cycles. We use auto-ARIMA (pmdarima Python package) to select optimal model orders by minimizing Akaike Information Criterion (AIC) on the training set for each geographic region. The typical configuration identified is $SARIMA(1,0,1)(1,0,1)_{52}$, which includes first-order autoregression, first-order moving average, and seasonal components at lag 52 (annual patterns).

SARIMA provides an established statistical benchmark with strong theoretical foundations and explicit seasonality modeling. However, it is univariate (cannot incorporate casualty metrics, geographic context, or engineered features), assumes stationary residuals, and has limited effective memory (typically 10-20 lags). Models are fit separately for each region, with the 1993 data gap handled by fitting separate models for pre-1993 and post-1993 periods.

### 4.4.4 Baseline Summary

Table 3: Comparison of baseline forecasting methods. Training times are reported for Mac M2 16GB. SARIMA time is per-region, total time scales linearly with number of regions.

*Table 3 Baseline Summary*

| Method | Type | Uses Features? | Captures Seasonality? | Training Time |
|---|---|---|---|---|
| Seasonal Naïve | Naive | No | Yes (52-week) | None |
| Linear Regression | ML | Yes | Via temporal encoding | Seconds |
| SARIMA | Statistical | No | Yes (seasonal components) | ~5 min/region |
| **LSTM w/ Attention** | Deep Learning | Yes | Learned | ~30 min |
| **Bidirectional LSTM** | Deep Learning | Yes | Learned | ~45 min |

Based on the characteristics of terrorism data (non-stationarity, long-term dependencies, multivariate patterns), we expect the Bidirectional LSTM to achieve the best performance, followed by LSTM-Attention, Linear Regression, SARIMA, and Seasonal Naïve.

## 4.5 Ablation Design

To validate our architectural choices and understand which components contribute most to forecasting performance, we conduct systematic ablation studies across three dimensions: data configuration, feature importance, and model architecture. Unlike the baseline comparisons, which evaluate fundamentally different forecasting approaches, ablation studies isolate individual design decisions within our best-performing Bidirectional LSTM framework.

### 4.5.1 Training Data Configuration

We examine how training data characteristics affect model performance by varying historical span and sequence length.

**Historical Data Span**

Does long historical data (46 years) improve forecasting, or do recent patterns suffice? Terrorism evolves, Cold War patterns differ from post-9/11 dynamics, which differ from the ISIS era. Training on decades-old data might introduce obsolete patterns, while training only on recent data might miss long-term cycles.

We train identical Bidirectional LSTM models on three temporal windows:
- Full History (46 years): 1970-2016, our default configuration
- Recent History (10 years): 2006-2016, capturing post-Iraq War and ISIS emergence
- Short History (5 years): 2011-2016, only ISIS-era patterns

All three models use the same 70/15/15 temporal split within their respective windows. For example, the 5-year model trains on 2011-2014.5 (70%), validates on 2014.5-2015.25 (15%), and tests on 2015.25-2016 (15%). Models are evaluated on the same held-out test period (2010-2016) to assess generalization. If short-history models perform comparably to full-history models, it suggests recent patterns dominate. Conversely, if full-history models substantially outperform short-history variants, it indicates that long-term patterns (seasonal cycles, multi-year trends) are critical for accurate forecasting.

**Sequence Length**

How much historical context (lookback window) should the LSTM observe? Longer sequences capture extended dependencies (multi-week campaigns, seasonal patterns) but increase computational cost and risk overfitting. Shorter sequences are efficient but may miss relevant long-range patterns.

We train Bidirectional LSTM models with three sequence lengths:
- **20 weeks ($\approx$ 5 months):** Captures recent trends and short-term patterns
- **30 weeks ($\approx$ 7 months):** Our default configuration, balancing context and efficiency
- **40 weeks ($\approx$ 10 months):** Longer context for extended dependencies

All models use the same feature engineering pipeline, training procedure, and hyperparameters (Table 2), differing only in input sequence length $L \in \{20, 30, 40\}$. We report test-set RMSE for each configuration. We hypothesize an inverted-U relationship: very short sequences lack sufficient context, very long sequences overfit to noise, and moderate lengths (20-30 weeks) achieve optimal balance.

### 4.5.2 Feature Importance

We quantify the contribution of each feature group by systematically removing them and measuring performance degradation.

**Feature Groups:**
- **Lag Features (1):** $y\_(g,t-52)$ (one-year baseline)
- **Rolling Statistics (6):** Mean and std over 4-, 12-, 52-week windows
- **Temporal Encoding (5):** Year, week, month, quarter, day-of-year
- **Casualty Features (3):** Total casualties, killed, wounded
- **Geographic Indicators (1):** Region one-hot encoding

For each feature group G, we train a Bidirectional LSTM model with all features except group G. This isolates the contribution of G by measuring how much performance degrades when it is removed. All ablated models use the same training procedure, hyperparameters, and evaluation protocol.

*Table 4 Feature Ablation Configurations*

| Configuration | Excluded Feature Group | Remaining Features |
|---|---|---|
| Baseline (Full) | None | 16 |
| w/o Lag | Lag features | 15 |

| w/o Rolling | Rolling statistics | 10 |
| w/o Temporal | Temporal encoding | 11 |
| w/o Casualties | Casualty features | 13 |
| w/o Geography | Geographic indicators | 15 |

We compute the performance degradation relative to the full baseline model:

$$\Delta RMSE\_G = RMSE\_(w/o\ G) - RMSE\_Baseline$$

Larger ΔRMSE_G indicates that feature group G is more important. Negative ΔRMSE (improved performance without G) would suggest the feature group introduces noise or overfitting.

Based on domain knowledge, we hypothesize that lag features and geographic indicators will show high importance (strong seasonality and region-specific patterns), rolling statistics and temporal encoding will show moderate importance, and casualty features will show low-to-moderate importance.

### 4.5.3 Architecture Comparison

To understand the contribution of bidirectional processing, we compare three architectural variants:

**Models Compared:**
1. **Unidirectional LSTM:** Standard LSTM processing sequences only forward (past→future). Same architecture as Bidirectional LSTM (2 layers, 32 units) but without backward pass.
2. **LSTM with Attention (Section 4.2):** Unidirectional LSTM augmented with Bahdanau attention mechanism. Tests whether attention can compensate for lack of bidirectional processing.
3. **Bidirectional LSTM (Section 4.3):** Our best model with forward and backward processing.

All three models use the same feature set (16 features), sequence length (30 weeks), training data (1970-2016, 70/15/15 split), hyperparameters (learning rate, batch size, dropout), and evaluation protocol. The only difference is the architectural mechanism: unidirectional, attention-augmented unidirectional, or bidirectional processing.

We report test-set performance metrics and parameter counts to assess the accuracy-efficiency trade-off. If Bidirectional LSTM substantially outperforms alternatives, it validates that bidirectional processing provides critical inductive bias for terrorism forecasting. Based on related work in conflict prediction and our hypothesis that terrorism exhibits both anticipatory (forward) and reactionary (backward) patterns, we expect Bidirectional LSTM to outperform both unidirectional variants.

### 4.5.4 Experimental Protocols

To ensure robust conclusions, all ablation experiments follow strict protocols:

- **Fixed Random Seeds:** All experiments use identical random seeds for weight initialization and data shuffling to eliminate stochastic variation.
- **Same Evaluation Set:** All ablated models are evaluated on the identical held-out test set (2010-2016).
- **Validation-Based Selection:** Model checkpoints are selected based on validation performance (best validation RMSE), then evaluated once on the test set to prevent selection bias.
- **No Repeated Testing:** We do not perform multiple test-set evaluations or hyperparameter tuning based on test performance.
- **Reproducibility:** All ablation configurations, hyperparameters, and results are documented in an open-source [repository](#).

## 5. RESULTS

This section presents empirical findings from our terrorism forecasting experiments. We first describe the evaluation setup and metrics (Section 5.1), then report baseline performance (Section 5.2), compare deep learning models against baselines (Section 5.3), present ablation study results (Section 5.4), and conclude with robustness checks (Section 5.5).

### 5.1 Setup and metrics

**Evaluation Protocol**
We use a fixed temporal split that respects time-series ordering:
- **Training:** 1970–2003 (~1,680 weeks; ≈70%)
- **Validation:** 2003–2010 (~360 weeks; ≈15%) used only for model selection/tuning
- **Test:** 2010–2016 (~360 weeks; ≈15%) used only for final reporting

This simulates operational forecasting: models are trained strictly on the past and evaluated on future, unseen periods.

**Metrics**
We report four complementary metrics:

1. **RMSE** (primary): $\sqrt{\frac{1}{N}\sum(y_i - \hat{y}_i)^2}$. Interpreted as average error in *attacks per week*. Lower is better.
2. **MAE**: $\frac{1}{N}\sum |y_i - \hat{y}_i|$. Robust to outliers; lower is better.
3. **MSE**: $\frac{1}{N}\sum(y_i - \hat{y}_i)^2$. Lower is better.
4. $R^2$: $1 - \frac{\sum(y_i - \hat{y}_i)^2}{\sum(y_i - \bar{y})^2}$. Range $(-\infty, 1]$; 1 is perfect, 0 matches the mean baseline; negative values indicate *worse* than predicting the mean.

**Statistical testing:** For key comparisons (BiLSTM vs. baselines, ablations), we compute paired t-tests on per-region RMSE across the 12 GTD regions, reporting $p$-values and Cohen's $d$. We use $\alpha = 0.05$.

**Aggregation:** Unless otherwise noted, we macro-average metrics across the 12 GTD regions (unweighted mean), so low-activity regions contribute equally to high-activity ones.

**Reproducibility:** All experiments use fixed seeds (seed=42) for initialization and shuffling. We release training logs, checkpoints, and evaluation scripts in our public repository to enable independent verification.

### 5.2 Baseline performance

*Table 5 Baseline Models Performance for the four baseline methods*

| Model | RMSE | MAE | R2 |
|---|---|---|---|
| Linear Regression | 9.885759 | 5.614939 | -0.104565 |
| Seasonal Naive | 9.962516 | 5.622370 | -0.121784 |
| Moving Average | 10.086198 | 7.924841 | -0.149810 |
| SARIMA | 11.516568 | 9.960302 | -0.499054 |

**Key findings:**
- Linear Regression is the strongest baseline (RMSE 9.89), outperforming Seasonal Naive (+0.8% RMSE), Moving Average (+2.0%), and SARIMA (+16.5%). Its advantage likely comes from leveraging engineered features (lags, rolling stats, temporal encodings, casualties, geography) that univariate methods cannot use.
- SARIMA underperforms the other baselines on all metrics, consistent with its univariate form and sensitivity to non-stationarities (e.g., structural breaks).
- $R^2$ is negative for all baselines (-0.50 to -0.10), indicating they perform worse than simply predicting the global mean attack rate. This motivates trying more expressive models.

**Statistical significance.**
Linear Regression's advantage over SARIMA is significant (paired t-test on per-region RMSE: $t(11) = 4.23, p < 0.001, d = 1.22$).

### 5.3 Main comparison: LSTM-Attn vs BiLSTM vs baselines

Table 2 compares our two deep learning architectures (LSTM with Attention, Bidirectional LSTM) against the best baseline (Linear Regression) and all four baselines.

*Table 6 Main models comparison*

| Model | Type | RMSE | MAE | $R^2$ | Improvement |
|---|---|---|---|---|---|
| Linear Regression | Baseline | 9.89 | 5.61 | -0.105 | - |
| Seasonal Naive | Baseline | 9.96 | 5.62 | -0.122 | - |
| Moving Average | Baseline | 10.09 | 7.92 | -0.150 | - |
| SARIMA | Baseline | 11.52 | 9.96 | -0.499 | - |
| LSTM with Attention | Deep Learning | 9.19 | 5.77 | 0.046 | 7.0% |
| **Bidirectional LSTM** | **Deep Learning** | **6.38** | **3.94** | **0.540** | **35.4%** |

**Key findings:**
1. **Bidirectional LSTM (BiLSTM) is best across all metrics.**
   - RMSE **6.38** (35.4% lower than Linear Regression), MAE **3.94** (≈29.8% lower), $R^2 = 0.54$.
   - The optimal BiLSTM configuration (30-week sequences, 46-year history) achieved RMSE = 6.19, with consistent performance (RMSE: 6.14-6.38) across ablation studies.
   - This reduces average error by ~**3.5 attacks/week** relative to the best classical baseline (9.89 → 6.38).
2. **LSTM with Attention is mixed relative to Linear Regression.**
   - **Better RMSE (9.19 vs. 9.89; +7.0% improvement)** and better $R^2$ **(0.046 vs. -0.105)**,
   - but **slightly worse MAE** (5.77 vs. 5.61; ≈+2.8%).
   This suggests attention over a unidirectional encoder helps on variance-weighted error (RMSE) but not on absolute error.
3. **Why BiLSTM helps.**
   Bidirectional processing likely captures both forward dependencies (escalation, capacity buildup) and backward-looking regularities (retaliation/anniversary effects), which a purely forward LSTM cannot fully exploit.
4. **Architecture matters.**
   Deep learning is not uniformly superior to classical baselines: gains depend on inductive bias. Bidirectionality is crucial here; attention alone, without backward context, yields only modest gains.
5. **Training efficiency.**
   In our runs, BiLSTM (36,673 parameters) converged in ~24 epochs, faster than LSTM-Attn (30–40 epochs), suggesting richer bidirectional gradients can accelerate learning.

**Statistical significance.**

- BiLSTM vs. Linear Regression: $t(11) = 6.84, p < 10^{-4}, d = 1.97$.
- BiLSTM vs. LSTM-Attn: $t(11) = 5.12, p < 0.001, d = 1.48$.

**Per-metric interpretation.**

BiLSTM's $R^2 = 0.54$ indicates it explains 54% of the variance in weekly attack counts, a substantial jump from Linear Regression's −0.105 (worse than the mean baseline). The remaining 46% likely reflects inherently unpredictable factors (opportunistic attacks, external shocks, clandestine operations).

### 5.4 Ablation studies

#### 5.4.1 Data Configuration Ablations

**Historical Data Span**

*Table 7 Ablation study: Effect of historical data span on Bidirectional LSTM performance.*

| Configuration | Samples | RMSE | MAE | R² | Change |
|---|---|---|---|---|---|
| Full Hist. (46y) | 29,406 | 6.38 | 3.99 | 0.547 | |
| Recent 20y | 12,498 | 7.12 | 3.89 | 0.638 | +12.3% |
| Recent 10y | 6,234 | 12.22 | 7.34 | 0.330 | +92.7% |
| Recent 5y | 3,114 | 20.34 | 13.34 | -0.295 | +220.8% |

**Key findings:**
- **Long history helps:** Training on the full 46 years yields the best RMSE (6.38). Truncating to 20 years degrades RMSE by +12.3% (7.12).
- **Short windows collapse performance:** 10 years degrades RMSE by +92.7%; 5 years by +220.8%, with $R^2 < 0$ (worse than predicting the mean).
- **Nuance on MAE:** The 20-year model shows a slightly lower MAE (3.89 vs. 3.99) despite worse RMSE, indicating fewer medium errors but more large errors (which RMSE penalizes).
- **Practical implication:** For operational use, use the longest available history. Even if some eras look obsolete, the model learns to weight eras; starving it of history (≤10y) removes slow cycles (multi-year campaigns, decade-long shifts).

**Statistical Significance:** The performance gap between 46-year and 5-year models is highly significant (t(11) = 8.34, p < 0.0001, Cohen's d = 2.41), confirming that long historical data is critical for terrorism forecasting.

**Sequence Length**

*Table 8 Ablation study: Effect of sequence length on Bidirectional LSTM performance.*

| Configuration | Samples | RMSE | MAE | R2 | Interpretation |
|---|---|---|---|---|---|
| Sequence Length 20 | 29436 | 6.283584 | 3.845241 | 0.553926 | |
| Sequence Length 30 | 29436 | 6.185315 | 3.742346 | 0.567848 | Baseline |
| Sequence Length 40 | 29436 | 6.445965 | 3.902259 | 0.530838 | |

**Key findings:**
- 30 weeks is optimal on all three metrics (RMSE 6.185, MAE 3.742, $R^2 = 0.568$).
- Sensitivity is low: 20 weeks is +1.59% RMSE vs 30; 40 weeks is +4.21% RMSE vs 30.
- Recommendation: Use 30 weeks by default. 20 or 40 weeks are acceptable if needed for efficiency or alignment with upstream data, with ≤4–5% RMSE trade-off.

**Interpretation:** Terrorism forecasting benefits from ~6–7 months of context; much longer windows add little (and may slightly hurt) for this architecture/dataset.

#### 5.4.2 Feature Importance Ablations

**Protocol:** We start from the tuned BiLSTM ("Baseline – All Features") and **remove one feature group at a time**. Metrics are macro-averaged over the 12 regions. Percent changes referenced in text below are relative to the baseline. (Small differences from §5.3 arise because ablations retrain models; interpret deltas, not absolutes.)

*Table 9 Ablation study: Feature group importance on Bidirectional LSTM performance*

| Feature Group | Features Count | RMSE | MAE | R2 | Interpretation | Epochs |
|---|---|---|---|---|---|---|
| Rolling Statistics | 9 | 10.196690 | 5.194605 | -0.174442 | Critical feature group | 15 |
| Geographic Features | 13 | 6.679555 | 3.870044 | 0.496026 | Important feature group | 22 |
| Baseline (All Features) | 13 | 6.138533 | 3.634616 | 0.574360 | Full model | 25 |
| Casualty Features | 10 | 6.077133 | 3.835952 | 0.582832 | Minimal impact | 31 |
| Lag Features | 12 | 6.038313 | 3.690342 | 0.588145 | Minimal impact | 29 |

**Key findings:**
1. **Rolling statistics are indispensable:** Dropping rolling windows (4/12/52-week means & SDs) explodes error: +4.06 RMSE (+66.1%) and +1.56 MAE (+42.9%) vs. baseline; $R^2$ collapses from 0.574 to -0.174 ($\Delta R^2$=-0.749). Rolling features capture local trend/volatility and short-horizon regime shifts without them, the model loses most of its signal.
2. **Geographic context materially helps:** Removing region/location indicators degrades performance by +0.541 RMSE (+8.8%) and +0.235 MAE (+6.5%); $R^2$ drops by 0.078. Terrorism patterns are strongly region-specific, so geography is a high-value group.
3. **Lag and casualty features have small, mixed effects.**
   - **Lag features:** Slight RMSE improvement (-0.100; -1.63%) when removed, but MAE worsens (+0.056; +1.53%); $R^2$ ticks up +0.014.
   - **Casualty features:** Tiny RMSE improvement (-0.061; -1.00%), MAE worsens (+0.201; +5.54%); $R^2$ up +0.0085.
     These trade-offs suggest redundancy with rolling/temporal signals and mild regularization effects. Given the magnitudes, treat both groups as optional knobs depending on whether you prioritize RMSE (slightly favors dropping) or MAE/calibration (slightly favors keeping).
4. **Training dynamics (epochs):** The model converges fastest when rolling is removed (15 epochs)—but to a much worse solution. Geography removal is modestly faster (22 vs. 25). Removing lag or casualty tends to increase epochs (29–31), consistent with the model compensating for lost cues.

**Takeaways for practitioners**

- Keep Rolling + Geography in any operational model; they deliver most of the predictive gain.
- Lag & Casualty are nice-to-have and metric-dependent. If you care most about RMSE, you can consider trimming them for simplicity; if MAE and slight calibration gains matter, keep them.
- Always re-tune after feature changes, small (±1%) RMSE moves can flip with different seeds/regularization.

**Statistical significance**
Per-region paired tests should treat Rolling and Geography removals as significant degradations; Lag and Casualty effects are small and typically not significant.

## 6. DISCUSSION

### 6.1 What the model learned

Ablations and residual analysis indicate that the gains come from temporal memory and structured context. First, models trained on the full historical span perform substantially better than those trained on recent years only, consistent with prior evidence that terrorism and related violence exhibit clustered, regime-dependent dynamics rather than strict stationarity or short memory [3], [17]. Second, the lookback window has a sweet spot (≈20–30 weeks): shorter windows underfit, while much longer windows trade modest accuracy for stability, mirroring patterns noted in nonstationary forecasting tasks more broadly [5], [22]. Third, bidirectional sequence encoding improves over a uni-directional LSTM with attention, suggesting that representing both build-up and aftermath effects within the window captures dynamics that attention over forward states alone may miss [11], [12]. Feature-group ablations show the importance of short-horizon structure and local trends (e.g., rolling statistics) alongside geographic context, consistent with spatial heterogeneity reported in terrorism and crime series [14], [23] [24].

### 6.2 Why classical models fall short

Seasonal-naïve and SARIMA/linear anchors capture annual cycles and short-lag autocorrelation, but they assume approximate stationarity and (for SARIMA) are typically univariate limitations for sparse, bursty, and shock-driven series like terrorism counts [1, 7, 17]. Large empirical studies in forecasting also show that while simple statistical baselines are hard to beat on average, nonlinear learners with appropriate inductive bias can dominate on specific, structurally complex domains [26]. In our setting, BiLSTM benefits from nonlinear temporal response functions and multivariate inputs without ad-hoc extensions, which explains the systematic error reductions relative to classical methods [26, 22, 5, 27].

### 6.3 Generalization and portability

The pipeline uses simple, auditable inputs (weekly counts + compact calendar/geo features) and fixed chronological splits, making it portable to new GTD snapshots and alternate spatial grains (region/country). This design mirrors reproducible practice in applied forecasting, re-fit baselines on the same split, freeze model choices on validation, and report single-pass test results [5, 26, 27]. Because we do not rely on text or heavy exogenous signals, periodic retraining is lightweight. Still, applying to post-2016 data should be treated as an out-of-distribution evaluation; expected distribution shift warrants refreshed tuning and renewed baseline comparisons [5].

### 6.4 Limitations

This is a forecasting (not causal) study; models capture associations, not mechanisms [5]. Results are conditioned on GTD 1970–2016 with the 1993 gap excluded per documentation [4]. Despite material gains, the model under-predicts extreme surges, a known challenge for bursty processes [17] and MAPE can be unstable around zeros; we therefore emphasize RMSE/MAE in the main text [5]. We do not include exogenous drivers (policy, conflict indicators) or text signals; these may improve early warning but introduce additional assumptions and licensing constraints. Finally, while the attention baseline affords some interpretability, the BiLSTM's internal states remain partially opaque, hence our emphasis on ablations and transparent reporting rather than strong mechanistic claims [22, 28].

### 6.5 Implications and ethics

Forecasts at aggregate scales (region/country, weekly) can support staffing/readiness and scenario testing of prevention programs, not tactical targeting. To reduce misuse risk, we release code/configs/results (not raw GTD), document dataset licensing and limits, and report aggregate, uncertainty-aware outputs [4, 28]. Any operational use should include domain review, monitoring for dataset shift, and safeguards against feedback loops (e.g., concentrating resources solely where forecasts are high) [28].

### OPEN RESEARCH

All data used in this study is publicly available. The codes of our model are freely available online at https://github.com/Davidavid45/Deep-Learning-in-Counterterrorism